\pdfoutput=1
\documentclass[11pt]{article}

\usepackage[final]{coling}

\usepackage{times}
\usepackage{latexsym}

\usepackage[T1]{fontenc}
\usepackage[utf8]{inputenc}

\usepackage{microtype}

\usepackage{inconsolata}

\usepackage{graphicx}

\usepackage{booktabs}

\usepackage{amsmath}
\usepackage{amssymb}
\usepackage{amsthm}
\usepackage[english]{babel}
\usepackage{makecell}
\usepackage{algorithm}
\usepackage{algorithmic}
\usepackage{multirow} 
\title{Parameter-Efficient Fine-Tuning of Large Language Models via Deconvolution in Subspace}

\author{
 \textbf{Jia-Chen Zhang\textsuperscript{1}},
 \textbf{Yu-Jie Xiong\textsuperscript{1}\thanks{Corresponding author.}},
 \textbf{Chun-Ming Xia\textsuperscript{1}},
 \textbf{Dong-Hai Zhu\textsuperscript{1}},
 \textbf{Xi-He Qiu\textsuperscript{1}}
\\
 \textsuperscript{1}School of Electronic and Electrical Engineering, Shanghai University of Engineering Science, \\
 333 Longteng Road, Songjiang District, Shanghai, China
\\
 \small{
   \textbf{Correspondence:} \href{mailto:xiong@sues.edu.cn}{xiong@sues.edu.cn}
 }
}

\begin{document}
\maketitle
\begin{abstract}
Large language model (LLM) is considered a milestone towards achieving Artificial General Intelligence (AGI). With its advanced emergent capabilities, it adapt to a wide range of specific applications. Fine-tuning LLMs for various downstream tasks has become a new paradigm. Low-Rank Adaptation (LoRA) is well-known for its parameter efficiency. It can reduce the number of parameters needed to fine-tune LLMs by several orders of magnitude. However, LoRA-based approaches encounter a significant limitation due to the bottleneck imposed by rank one decomposition. As the parameters count in LLMs increase, even rank one decomposition might surpass the number of parameters truly necessary for handling more downstream tasks.
In this paper, we propose a new method for Parameter-Efficient Fine-Tuning (PEFT) via deconvolution in subspace, dubbed as DCFT. 
We innovatively use deconvolution to complete details and enhance knowledge in subspace incremental matrices, and dynamically control parameters by adjusting the kernel size, unconstrained by rank-one decomposition.
Extensive experiments are conducted to validate the effectiveness of DCFT. Results show that compared to LoRA, DCFT achieve an 8$\times$ reduction in parameters, and still achieves highly impressive performance. Our code is available here: \url{https://github.com/Godz-z/DCFT}.
\end{abstract}

\section{Introduction}
LLMs are considered a potential spark for AGI \cite{xi2023risepotentiallargelanguage}. Due to their excellent situational adaptability and language comprehension abilities, LLMs have become the cornerstone of NLP tasks \cite{devlin-etal-2019-bert,zhuang-etal-2021-robustly,he2021deberta,Radford2019LanguageMA}. The success of GPT-3.5 has mainstreamed the development of LLMs towards larger parameter counts \cite{10.5555/3600270.3602281}. Over the past few years, the parameter scale of pretrained language models has increased by thousands of times; for example, PaLM \cite{10.5555/3648699.3648939} contains up to 540 billion parameters, while GPT-4 \cite{openai2023gpt4} contains up to 100 trillion parameters. Nevertheless, due to the knowledge boundaries of LLMs, their abilities in some downstream tasks are still limited. To expand these knowledge boundaries, it remains necessary to fine-tune LLMs on downstream tasks \cite{Qiu_Sun_Xu_Shao_Dai_Huang_2020,zhuang-etal-2021-robustly}.

However, the time and resource costs required to fine-tune all parameters are prohibitive. Various methods for parameter-efficient fine-tuning have been proposed to reduce these costs. The PEFT methods are divided into two categories based on whether the pretrained parameters are frozen \cite{lialin2023scalingscaleupguide}. Considering the potential issues of catastrophic forgetting and reduced generalization ability that may arise from modifying model parameters and architecture, LoRA \cite{hu2022LoRA} has become one of the most widely applied PEFT methods.
LoRA introduces incremental updates to pretrained weights via the product of two low-rank matrices. It reduces training overhead by up to 70\% and achieves comparable or superior performance to fine-tuning.

\begin{figure*}
\centerline{\includegraphics[width=\textwidth]{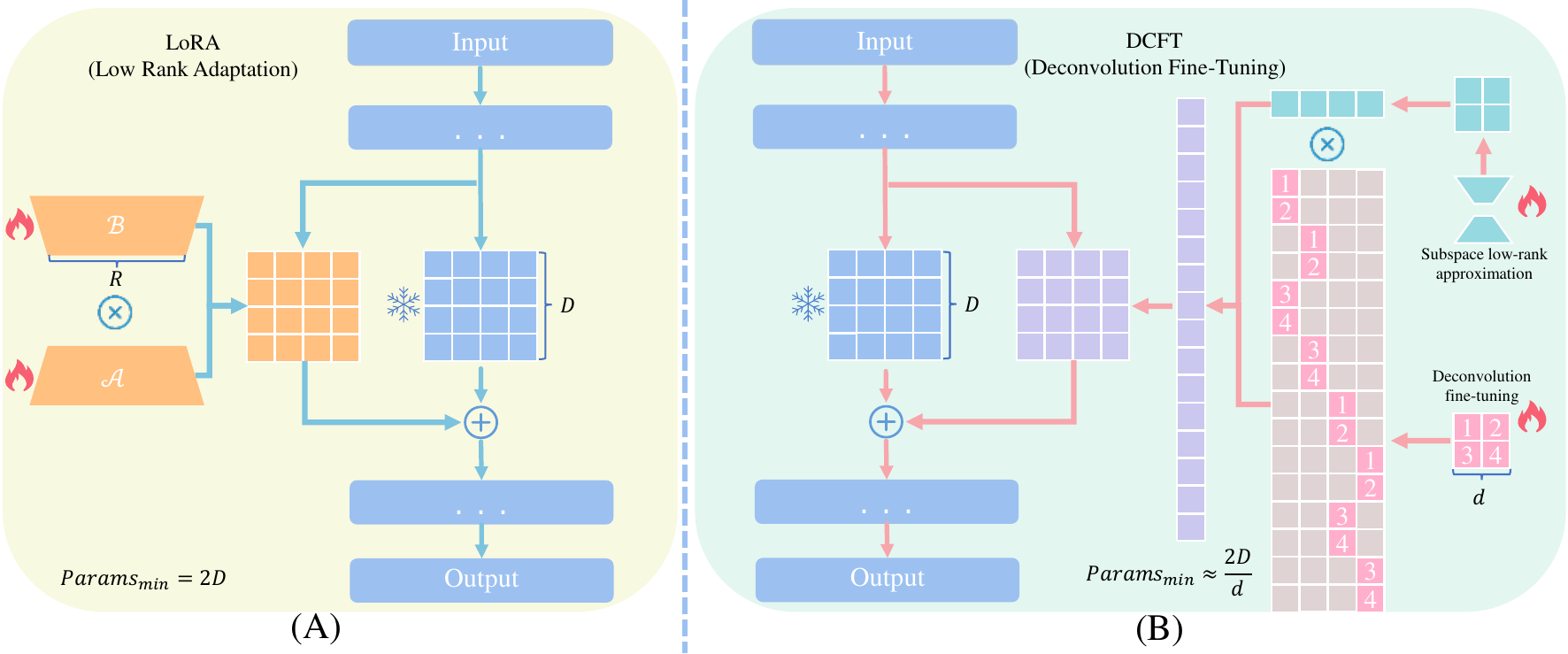}}
  \caption{An illustration of the differences between LoRA and DCFT. The parameter calculation results represent the model's parameters when $r = 1$. $D$ represents the dimension of the pretrained weights, and $d$ represents the dimension of the convolution kernel.}
  \label{photomain}
\end{figure*}
Nonetheless, due to the constraints of rank-one decomposition, LoRA-based approaches \cite{zhang2023adaptive,ding-etal-2023-sparse,liu2024dora} have limitations in parameter efficiency. With the increasing parameter counts in LLMs, even rank one decomposition may exceed the necessary number of parameters for managing more complex models. 
Many researchers are dedicated to breaking this limitation. LoRA-FA \cite{zhang2023lorafamemoryefficientlowrankadaptation} reduces trainable parameters and memory further by freezing the matrix $\mathcal{A}$ in LoRA. NOLA \cite{koohpayegani2024nola} decomposes LoRA's matrix $\mathcal{A}$ and $\mathcal{B}$ further into several small random matrices. Yet, these methods consistently follow the fundamental architecture of LoRA, encountering the persistent issue that parameter variations are contingent upon the size of $R$. Additionally, they have frozen some parameters, significantly constraining the model's capacity to learn from new data.

In this paper, we innovatively combine the feature reconstruction ability of deconvolution and the efficiency of subspace learning, dubbed as Deconvolution Fine-Tuning (DCFT).
Deconvolution is a kind of CNN that performs upsampling by transposing the convolution kernel matrix \cite{5539957}. It is widely used in the fields of image processing and computer vision, commonly applied to tasks such as image reconstruction, semantic segmentation, and feature enhancement \cite{dumoulin2018guideconvolutionarithmeticdeep,10.1007/978-3-319-10590-1_53}. For the first time, we apply deconvolution to the fine-tuning of large models, restoring and enhancing features learned from incremental matrices through deconvolution.
We first learn a set of low-rank incremental matrices for LLMs within the subspace. Subsequently, by integrating orthogonal projection theory, we apply orthogonal constraints to maximize the learning space of these matrices. 
Then, we enhance and complete the matrices learned in the subspace through deconvolution to adapt them to the dimensions of the incremental matrices. By setting the stride equal to the convolution kernel size, we simplify the computation while better preserving the knowledge learned in the subspace and enhancing the stability of models.

Extensive experiments are conducted on various tasks and models to demonstrate the effectiveness of DCFT. Specifically, natural language understanding GLUE \cite{wang-etal-2018-GLUE} is evaluated using DeBERTa and RoBERTa \cite{he2021deberta, he2023debertav, zhuang-etal-2021-robustly}. The findings, including qualitative and quantitative results, indicate that DCFT outperforms existing methods. The contributions of this paper are as follows:
\begin{itemize}
\item[$\bullet$] We propose Deconvolution Fine-Tuning (DCFT), a novel method employing deconvolution to reconstruct subspace features in incremental matrices. Compared to LoRA, DCFT achieves improved performance while using only 14\% of the parameters.
\end{itemize}
\begin{itemize}
\item[$\bullet$] We significantly reduce the computational complexity of DCFT and improve the fine-tuning efficiency through a series of methods including low-rank matrices, equal kernel stride, and larger convolution kernels.
\end{itemize}
\begin{itemize}
\item[$\bullet$]Extensive experiments are conducted to validate the effectiveness of our method. Notably, our model consistently surpasses parameter-efficient baselines, achieving superior performance with fewer parameters across a broad spectrum of downstream tasks.
\end{itemize}

\section{Related Work}
\subsection{Parameter-Efficient Fine-Tuning}
PEFT is a technique for adapting LLMs by optimizing a small set of parameters while keeping most of the pre-trained parameters unchanged. This approach effectively reduces memory and computational costs, allowing large models to be applied in resource-constrained environments. Common PEFT techniques include Low-Rank Adaptation (LoRA) \cite{hu2022LoRA}, Mixture of Experts (MoE), and Adaptable Embeddings (Adapters) \cite{pmlr-v97-houlsby19a,li-liang-2021-prefix}. These methods enhance the performance of the model on specific tasks by updating only a small part of the model while keeping the majority of the weights fixed, thus preserving the pre-trained knowledge.

\subsection{LoRA-based approach}
Due to the high cost of full-parameter fine-tuning of LLMs, numerous parameter-efficient methods have been introduced. Among these, LoRA \cite{hu2022LoRA} stands out by updating only a small subset of the model's parameters, thereby reducing memory overhead while maintaining performance on par with full-parameter fine-tuning. As Equation \ref{eq:1} shows, LoRA adds a low-rank adapter to the frozen pre-trained weights to learn incremental matrices.
\begin{equation}
  \label{eq:1}
W = W^{(0)} + \Delta  = W^{(0)} + \mathcal{BA},
\end{equation}
where $\Delta  \in  \mathbb{R}^{din\times dout}$, $\mathcal{A} \in  \mathbb{R}^{r\times dout}$, and $\mathcal{B} \in \mathbb{R}^{din\times r}$, with $r\in (din, dout)$. The dimensions of $din$ and $dout$ are the same as those of the pre-trained matrix $W$. During fine-tuning, only $A$ and $B$ are updated. The rank $r$ is chosen to be much smaller than the dimension of $W$. With less than 0.5\% additional trainable parameters, the training overhead can be reduced up to 70\%. 

Building on this, AdaLoRA \cite{zhang2023adaptive} introduces an adaptive mechanism for pruning, LoRA-FA \cite{zhang2023lorafamemoryefficientlowrankadaptation} freezes the parameters of matrix $\mathcal{A}$, and NoLA \cite{koohpayegani2024nola} further decomposes matrix $AB$ into the sum of multiple random matrices. DoRA \cite{liu2024dora} trains an additional set of parameters by decomposing the pre-trained weights. However, all these methods are based on the LoRA architecture, which leads to a significant issue: the number of trainable parameters is controlled by the rank $r$, which can only increase in multiples, and there exists a bottleneck in rank-one decomposition.

\subsection{Convolution and Deconvolution}
\textbf{Convolution} is a fundamental operation in many areas of signal processing and deep learning, especially in the context of Convolutional Neural Networks (CNNs) \cite{726791}. It involves sliding a filter (or kernel) over the input data to produce a feature map, which captures important spatial hierarchies in the data. Mathematically, the convolution operation for a 2D input can be expressed as:
\begin{equation}
  \label{eq:1}
Y(i, j) = \sum_{m} \sum_{n} X(i+m, j+n) \cdot W,
\end{equation}
where $Y(i,j)$ represents the output feature map, $X$ is the input feature map, $W$ is the filter or kernel. The operation essentially involves multiplying the filter with corresponding input values and summing them to obtain the output at each spatial position. Convolution is widely used for tasks such as image recognition, object detection, and segmentation due to its ability to extract meaningful features from raw data.

\textbf{Deconvolution} \cite{5539957}, also known as transposed convolution or upsampling, is the reverse operation of convolution. While convolution reduces the spatial dimensions of the input, deconvolution aims to increase them, making it a crucial operation in tasks requiring output with higher spatial resolution, such as image generation and semantic segmentation.
The deconvolution operation can be mathematically represented as:
\begin{equation}
  \label{eq:1}
X(i, j) = \sum_{m} \sum_{n} Y(i - s_m, j - s_n) \cdot W,
\end{equation}
where $X(i,j)$ is the reconstructed input, $Y$ is the output feature map from the previous layer, $W$ is the filter or kernel, and $s_{m}$ and $s_{n}$ denote the strides in the spatial dimensions. The transposed nature of this operation is reflected in the indices $i - s_{m}$ and $j - s_{n}$, where the input is effectively expanded.

Deconvolution is not merely the inversion of the convolution process but involves a more complex transformation where the spatial resolution of the feature map is increased, often leading to more refined and higher-quality outputs. This operation is critical in tasks where the generation or reconstruction of detailed spatial information is necessary.

\section{Our Method}
In this section, we provide a detailed description of the DCFT method's design, which enables flexible control over fine-tuning parameters and overcomes the constraints of rank one decomposition. Additionally, we introduce the optimization for efficiency employed in our method.

\subsection{Deconvolution Fine-Tuning}
The overall pipeline of DCFT is depicted in Figure~\mbox{\ref{photomain}}. We first train a set of low-rank matrices $\mathcal{A}$ and $\mathcal{B}$ in the subspace. Orthogonal constraints are applied to matrix $\mathcal{A}$ and $\mathcal{B}$ to maximize the learning space:
\begin{equation}
  \label{eq:2}
        R(\mathcal{A, B}) = \left\| \mathcal{A}^T \mathcal{A} - \mathcal{I} \right\|_{F}^{2} + \left\| \mathcal{B B}^T - \mathcal{I} \right\|_{F}^{2},
\end{equation}
where equation \ref{eq:2} enhances the orthogonality of matrix $\mathcal{A}$ and $\mathcal{B}$ by approximating $\mathcal{A}^T\mathcal{A} =\mathcal{B B}^T = \mathcal{I}$. 
Then, by applying the deconvolution operation, we upscale the knowledge learned in the subspaces, increasing the spatial dimensions of the incremental matrix. By leveraging the existing knowledge in the subspaces, we predict missing parts. This step allows us to obtain the incremental matrix with minimal parameters. Moreover, because deconvolution can learn the mapping from the subspace to the parent space, the incremental matrix obtained through deconvolution exhibits smoother and more precise details of implicit knowledge compared to that derived directly from the product of low-rank matrices.

As shown in Equation \ref{eq:3}, the sub-feature matrix $\mathcal{F}$, obtained by multiplying the low-rank matrices, is subsequently multiplied by the transpose of the convolution kernel matrix $ C_{s}^{T}$. This step completes and enhances the knowledge learned in the subspace. Ultimately, an incremental matrix of the same dimension as the frozen pretrained weights is obtained. The parameter increment matrix is added to the pretrained parameters to adapt to various downstream tasks, offering a plug-and-play capability. Our forward propagation process is as follows:
\begin{equation}
  \label{eq:3}
Conv^{T}(\mathcal{F}) = \mathcal{C}_{s}^{T} \cdot \mathcal{F} = \mathcal{C}_{s}^{T} \cdot (\mathcal{B} \cdot \mathcal{A}),
\end{equation}
\begin{equation}
  \label{eq:4}
W = W^{(0)} + \Delta = W^{(0)} + Conv^{T}(\mathcal{F}),
\end{equation}
where $\Delta  \in  \mathbb{R}^{(din\times dout)}$, $F \in  \mathbb{R}^{(\frac{din}{d}\times \frac{dout}{d})}$,  The dimensions of $din$ and $dout$ are the same as those of the pre-trained matrix $W$. $Conv^{T}(*)$ denotes the deconvolution operation. $s$ represents the stride of the transposed convolution. All of our trainable parameters are initialized randomly. We summarize the detailed algorithm in Algorithm~\mbox{\ref{alg:our_algorithm}}.

\begin{algorithm}[t!]
 	\caption{{Algorithm of DCFT}} 
 	\label{alg:our_algorithm}
 	\begin{algorithmic}[1]
 		\STATE {{\bfseries Input:} Dataset $ \mathcal{D} $;total step $T$; Kernel size $d$; stride $s$. }  
            \STATE Create matrix $ \mathcal{A}, \mathcal{B}, \mathcal{C}_{d} $;
 		\FOR{$ t = 1, \dots, T $}  
 		\STATE Sample a mini-batch from $\mathcal{D}$;\\
            \texttt{//Subspace low-rank approximation}
            \STATE Compute input matrix $\mathcal{F} = \mathcal{A} * \mathcal{B}$;\\
            \texttt{//Deconvolution fine-tuning}
            \STATE Set stride $s$ $=$ kernel size $d$;
 		\STATE Reshape matrix $\mathcal{C}_{d}$ $\rightarrow$ matrix $\mathcal{C}_{s}^{T}$;
            \STATE Deconvolution reconstruction $Conv^{T}(\mathcal{F})$;
 		\STATE Compute incremental matrix $\Delta$ as \ref{eq:4};
            \STATE Add to the pre-trained weights.
 		\ENDFOR 
            % \STATE \textbf{Output:}  { The fine-tuned parameters $(\Lambda, i)$.} 
	\end{algorithmic}
\end{algorithm}
\subsection{Optimization for Efficiency}
In this section, we introduce our approaches for optimizing computational efficiency. Compared to LoRA, which involves only two matrix multiplications, deconvolution operations significantly increase computational complexity. To address this challenge, we have implemented the following three principal measures to reduce the computational complexity of DCFT.

\subsubsection{Low-Rank Matrice}
We employed the LoRA methodology to decompose the initial matrix $F$ into the product of $\mathcal{A} * \mathcal{B}$. As shown in equation \ref{eq:3}, where $\mathcal{A} \in  \mathbb{R}^{(k\times \frac{dout}{d})}$ and $\mathcal{B} \in \mathbb{R}^{(\frac{din}{d}\times k)}$. In this paper, we set $k=1$. By using low-rank matrices, we have reduced the parameters by an order of magnitude, effectively enhancing parameter efficiency and training speed. The change in the number of parameters after adopting the low-rank matrix is shown in Equations \ref{eq:5} and~\mbox{\ref{eq:6}}.
\begin{equation}
  \label{eq:5}
        Params_{before} = ( \frac{d_{out}}{d} \times \frac{d_{out}}{d} ) \cdot layers,
\end{equation}
\begin{equation}
  \label{eq:6}
        Params_{after} =  ( \frac{d_{out}}{d} + \frac{d_{out}}{d} ) \cdot layers.
\end{equation}

\subsubsection{Equal Kernel Stride}
DCFT sets the stride equal to the dimension of the convolution kernel, ensuring that there is no overlap between the outputs of each convolution operation. This means that each input element is computed only once. This significantly reduces redundant calculations and prevents instability and unpredictability caused by overlapping computations. Under the condition that the output matrix dimensions are fixed, the Equation for the input matrix is as follows:
\begin{equation}
  \label{eq:7}
       D_{in} = \frac{(D_{out} - K + 2P)}{S} + 1,
\end{equation}
where K represents the kernel size, P represents the padding size, and S represents the stride. When K equals S, the equation simplifies to:
\begin{equation}
  \label{eq:8}
       D_{in} = \frac{D_{out} }{S}.
\end{equation}

Therefore, setting the stride equal to the dimension of the convolution kernel can effectively reduce the requirements for the input matrix, maximize the capability of deconvolution feature completion, and further reduce the computation and parameters. 

\subsubsection{Convolution kernel size}
DCFT can control the number of parameters by adjusting the size of the convolution kernel. When a larger convolution kernel $(d=8)$ is used, the network architecture is further simplified, leading to a more uniform distribution of computations, effectively eliminating checkerboard artifacts, and enhancing the quality of upsampling. Moreover, the adoption of a larger kernel significantly reduces the number of parameters in the deconvolution input matrix. This reduction in parameter gradient computations accelerates model convergence, thereby enhancing the fine-tuning efficiency of DCFT. 

When a smaller convolution kernel $(d=2)$ is used, the network is better able to capture the local features of the input data, allowing the model to better understand and extract local information, which aids in the model's generalization ability on new data. Detailed experimental validation is provided in Section \ref{size}.

\begin{table*}[!ht]
    \centering
    \scalebox{0.78}{
    \renewcommand\arraystretch{1.3}
    \begin{tabular}{l|l|c|ccccccccc}
    \toprule
        \textbf{Model}&\textbf{Method} & \textbf{\#Params} & \textbf{CoLA} & \textbf{SST-2} & \textbf{MRPC} & \textbf{QQP} & \textbf{STS-B} & \textbf{MNLI} & \textbf{QNLI} & \textbf{RTE} & \textbf{Avg.}  \\  
        \midrule
        DeB$^{V3}_{base}$&Fine-Tune  &184M  & 69.21 & 95.64 &89.22 & 91.10 & 91.59 &89.98/89.95 &93.78 &82.49 &87.82 \\ 
        \midrule
        \multirow{8}{*}{\shortstack{DeB$^{V3}_{base}$}} 
        &Bitfit &0.1M &68.70 &94.38 & 87.16 & 86.68 &89.71 &87.45/87.45 &91.90 & 76.12 &85.18 \\
        &HAdapter &0.31M &67.65 &95.41 & 89.25 & 90.18 &91.31 &\textbf{90.10}/90.02 &93.52 & 83.39 &87.60 \\
        &PAdapter &0.3M &69.06 &94.72 & 89.71 & 90.01 &91.38 &89.89/90.06 &93.87 & \underline{84.48} &87.90 \\
        &LoRA &0.17M &68.60 &94.95 &88.24 &89.79 &91.41 &\underline{90.09}/\underline{90.28}  &93.35 &81.29&87.23 \\
        &SoRA &0.12M &70.24 &95.14 &89.22 &90.13 &91.41 &90.08/\textbf{90.41} &93.43 &83.02&87.85 \\
        &LoRA-FA &0.12M &70.24 &95.14 &89.22 &90.13 &91.41 &90.08/\textbf{90.41} &93.43 &83.02&87.85 \\        
        &DCFT$^{\Delta}$(ours) &\textbf{0.024M} &\textbf{71.07} &\underline{95.53} &\underline{90.93} &89.20 &\underline{91.56} &89.40/89.63 &\underline{93.59} &\textbf{88.45} &\underline{88.73} \\
        &DCFT$^{\dagger}$(ours)  &\underline{0.079M} &\underline{70.43} &\textbf{96.10} &\textbf{91.42} &89.60 &\textbf{91.76} &89.97/90.12 &\textbf{94.14} &\textbf{88.45} &\textbf{88.99} \\
        \midrule
        \midrule
        RoB$_{large}$&Fine-Tune  &335M  & 68.0 & 96.4 &90.9 & 92.2 &92.4 &90.2 &94.7 &86.6 &88.93 \\
        \midrule
        \multirow{6}{*}{\shortstack{RoB$_{large}$}} 
        &HAdapter &0.8M &66.3 &96.3 &87.7 &\underline{91.5} &91.5&\underline{90.3}  &94.7 &72.9&86.40\\
        &PAdapter &0.8M &67.8 &\textbf{96.6} &89.7 &\textbf{91.7} &91.9&\textbf{90.5}  &\underline{94.8} &83.8&88.35\\
        &LoRA &0.8M &68.0 &96.2 &90.2 &91.1 &91.9&90.0  &94.4 &86.3&88.51\\
        &LoRA-FA &3.7M &68.0 &96.0 &90.0 &91.1 &\underline{92.0} &90.1  &94.4 &86.1 &88.46 \\
        &DCFT$^{\Delta}$(ours) &\textbf{0.06M} &\underline{68.5} &96.0 &\textbf{90.9} &90.6 &\textbf{92.2} &90.1  &94.2 &\underline{87.0} &\underline{88.69}\\
        &DCFT$^{\dagger}$(ours) &\underline{0.21M} &\textbf{69.3} &\underline{96.4} &\underline{90.7} &\underline{91.5} &\textbf{92.2} &\textbf{90.5}  &\textbf{95.1} &\textbf{89.9} &\textbf{89.31}\\
        % \midrule
        % \midrule
        % DeB$^{V2}_{xxl}$&Fine-Tune  &1.5B  & 72.0 & 97.2 &92.0 & 92.7 &92.9 &91.8 &96.0 &93.9 &91.06 \\
        % \midrule
        % \multirow{2}{*}{\shortstack{DeB$^{V2}_{xxl}$}} 
        % &LoRA &4.7M &72.4 &\textbf{96.9} &\textbf{92.6} &92.9 &93.0&\textbf{91.9} &96.0 &94.9&91.33\\
        % &DCFT$^{\Delta}$(ours) &\textbf{0.18M} &\textbf{73.0} &96.8 &91.4 &\textbf{93.0} &\textbf{93.1} &91.1  &\textbf{96.2} &\textbf{91.7} &\textbf{90.79}\\

    \bottomrule 
    \end{tabular}}
    \caption{Test results of DCFT and other baselines on the GLUE benchmark are presented. We report both matched and mismatched accuracies for MNLI, Matthew's correlation for CoLA, Pearson correlation for STS-B, and accuracy for other tasks. DCFT$^{\Delta}$ represents the most efficient results, while DCFT$^{\dagger}$ denotes the optimal results. Higher scores indicate better performance for all metrics. We employ consistent hyperparameters, detailed in Appendix \ref{A}. Optimal values are used as the final results. The best result is highlighted in \textbf{bold}, and the second best is \underline{underlined}.}
    \vspace{-0.4cm}
    \label{tab:main}
\end{table*}

\section{Experiments}
In this section, we conduct a comprehensive evaluation of our method through a series of experiments.
We implement DCFT for fine-tuning DeBERTaV3-base \cite{he2023debertav}, DeBERTaV2-xxl and RoBERTa-large \cite{zhuang-etal-2021-robustly}, we assess the effectiveness of the proposed algorithm on natural language understanding (NLU) and question answering (QA).
% we assess the effectiveness of the proposed algorithm on natural language understanding(NLU) using the GLUE benchmark \cite{wang-etal-2018-GLUE} and SQuAD \cite{rajpurkar-etal-2016-squad} benchmark on question answering(QA).
\subsection{Experimental Settings}
\textbf{Implementation Details.}  We implement all algorithms using PyTorch \cite{10.5555/3454287.3455008}. Our implementation is based on the publicly available code-base of Huggingface Transformers3 \cite{wolf-etal-2020-transformers}. Experiments involving the DeBERTaV3-base \cite{he2021deberta} are performed on an NVIDIA 2080-ti GPU and experiments about RoBERTa-large \cite{zhuang-etal-2021-robustly} and  DeBERTaV2-xxl are conducted on NVIDIA A800 GPU. 
Due to the bottleneck of rank one in the LoRA-base model, we compare the results of rank one with DCFT.\\
\noindent\textbf{Baselines.} Our baselines include full-parameter fine-tuning and other widely recognized parameter-efficient methods, including Bitfit \cite{zhang2022platon} , Adapter tuning \cite{pmlr-v97-houlsby19a,pfeiffer-etal-2021-adapterfusion}, LoRA \cite{hu2022LoRA}, AdaLoRA \cite{zhang2023adaptive}, LoRA-FA \cite{zhang2023lorafamemoryefficientlowrankadaptation}, and SoRA \cite{ding-etal-2023-sparse}.\\

\subsection{Natural Language Understanding}
\noindent\textbf{Models and Datasets.}  
We fine-tune DeBERTaV3-base, DeBERTaV2-xxl \cite{he2023debertav} and RoBERTa-large \cite{zhuang-etal-2021-robustly} and adopt the GLUE benchmark \cite{wang-etal-2018-GLUE} for evaluation. It's a widely recognized benchmark for natural language understanding, including CoLA \cite{warstadt-etal-2019-neural}, SST-2 \cite{socher-etal-2013-recursive}, MRPC \cite{dolan-brockett-2005-automatically}, QQP \cite{wang-etal-2018-GLUE}, STS-B \cite{wang-etal-2018-GLUE}, MNLI \cite{williams-etal-2018-broad}, QNLI \cite{rajpurkar-etal-2016-squad} and RTE \cite{10.1007/11736790_9,giampiccolo-etal-2007-third,Bentivogli2011TheSP}. Dataset details are summarized in Appendix \ref{A}.

\noindent\textbf{Main results}. We first evaluate the task of Natural Language Understanding based on the GLUE benchmark. Two results for DCFT are reported: the one with the highest training efficiency and the one with the strongest performance. The experimental performance of DCFT and other baselines is shown in Table \ref{tab:main}. To ensure the fairness of the experiments, we used the same hyperparameters and ensured that the parameters were closely matched. The results show that on the base model, DCFT outperforms the baselines in most tasks, and the trainable parameters are significantly fewer than those of the baselines. For example, in the RTE task, the accuracy of DCFT reached 88.45\%, which is 5.43\% higher than that of SoRA. This demonstrates that for some simple tasks, even if the parameters are set to $r=1$, the trainable parameters still far exceed the requirements.
On the RoBERTa-large and DeBERTaV2-xxl models, due to the expansion of the dimensionality of the incremental parameter matrix, the issue of parameter redundancy caused by rank-one decomposition is further enhanced. This results in more significant performance improvements for DCFT in most tasks. For instance, in the CoLA task, DCFT uses only 0.21M and 0.18M trainable parameters, leading LoRA by 1.3\% and 0.6\%, respectively, and surpassing other LoRA-based approaches.
\begin{table}[!ht]
    \centering
    \scalebox{0.72}{
    \renewcommand\arraystretch{1.2}
    \tabcolsep=0.2cm
    \begin{tabular}{l|c|cccc}
    \toprule
        \textbf{Method} & \textbf{\#Params} & \textbf{CoLA} & \textbf{SST-2}& \textbf{STS-B}& \textbf{QNLI}\\ 
        Fine-Tune  &1.5B  & 72.0 & 97.2 &92.9 &96.0 \\
        \midrule
        LoRA &4.7M &72.4 &\textbf{96.9} &93.0&96.0\\
        DCFT$^{\Delta}$(ours) &\textbf{0.18M} &\textbf{73.0} &96.8 &\textbf{93.1} &\textbf{96.2}\\
    \bottomrule 
    \end{tabular}}
    \caption{Results of DCFT fine-tuned on four different datasets on DeBERTaV2-xxl. Only the results with the best training efficiency are reported. The best result is highlighted in \textbf{bold}.}
    \vspace{-0.4cm}
    \label{QA}
\end{table}
\subsection{Question Answering}
\noindent\textbf{Models and Datasets.}  
We fine-tune DeBERTaV3-base \cite{he2023debertav} on two question answering (QA) datasets: SQuAD v1.1 \cite{rajpurkar-etal-2016-squad} and SQuAD v2.0 \cite{rajpurkar-etal-2018-know}. The Stanford Question Answering Dataset (SQuAD) is a key NLP resource featuring over 100,000 questions on Wikipedia articles for training question-answering models. Dataset details are summarized in Appendix \ref{B}.

\begin{table}[!ht]
    \centering
    \scalebox{0.72}{
    \renewcommand\arraystretch{1.2}
    \tabcolsep=0.2cm
    \begin{tabular}{l|l|c|cc}
    \toprule
        \textbf{Model}&\textbf{Method} & \textbf{\#Params} & \textbf{SQuADv1.1} & \textbf{SQuADv2.0}\\  
        \midrule
        DeB$^{V3}_{base}$&Fine-Tune  &184M  & 86.0 / 92.7 & 85.4 / 88.4  \\ 
        \midrule
        \multirow{4}{*}{\shortstack{DeB$^{V3}_{base}$}} 
        &HAdapter &0.08\% &84.4 / 91.5 &83.4 / 86.6 \\
        &PAdapter &0.08\% &84.4 / 91.7 &84.2 / 87.2 \\
        &LoRA &0.08\% &86.6 / 92.9 &83.6 / 86.7 \\
        &DCFT(ours)  &\textbf{0.04\%} &\textbf{87.3} / \textbf{93.4} &\textbf{84.7} / \textbf{87.7} \\
    \bottomrule 
    \end{tabular}}
    \caption{Test results of DCFT and other baselines on the SQuAD benchmark are presented. We report EM and F1. Higher scores indicate better performance for all metrics. We employ consistent hyperparameters, detailed in Appendix \ref{B}. The best result is highlighted in \textbf{bold}.}
    \vspace{-0.4cm}
    \label{QA}
\end{table}

\begin{table*}[!htbp]
\centering
\scalebox{0.85}{
\tabcolsep=0.4cm
\label{tab:summarization}
\begin{tabular}{c|c|c|c|c|c|c|c|c}
\toprule
 {\bf Method\&\# Param} & \textbf{Metric} & {\bf SST-2} & {\bf CoLA}& {\bf QNLI }& {\bf RTE}& {\bf MRPC}& {\bf STS-B}&
{\bf Avg.}\\
\midrule 
{\bf Full FT} & {Acc} & { 95.63} & {69.19} & {94.03} &{83.75}&{89.46}&{91.60}&{87.28}\\
\midrule 
{\bf DCFT$(d=2)$}& {Acc} & {\bf96.10} & {70.43} & {\bf94.14}& {\bf88.45}& {\bf91.42}& {91.76}& {\bf88.72}
\\
{\bf 0.079M} & {Time} & {9.57h} & {0.97h}& {13.67h}& {1.63h}& {1.25h}& {0.97h}& {4.68h}
\\
\midrule
{\bf DCFT$(d=4)$}& {Acc} & {95.76} & { 69.24}& {93.76}& {\bf88.45}& {90.20}& {\bf91.80}& {88.20}
\\
{\bf 0.041M} & {Time} & {7.32h} & {0.68h}& {12.27h}& {1.47h}& {1.27h}& {0.79h}& {3.97h}
\\
\midrule
{\bf DCFT$(d=6)$}& {Acc} & {95.07} & {70.69} & {93.65}& {86.28}& {89.95}& {91.73}& {87.90}
\\
{\bf 0.029M} & {Time} & {6.97h} & {0.64h}& {12.16h}& {1.45h}& {1.24h}& {0.77h}& {3.87h}
\\
\midrule
{\bf DCFT$(d=8)$}& {Acc} & {95.53} & {\bf 71.07}& {93.59}& {\bf88.45}& {90.93}& {91.56}& {88.52} 
\\
{\bf 0.024M} & {Time} & {\bf 5.01h} & {\bf0.54h}& {\bf10.19h}& {\bf1.08h}& {\bf0.98h}& {\bf0.63h}& {\bf3.16h}
\\
\midrule
{\bf DCFT$(d=12)$}& {Acc} & {95.76} & {67.87} & {93.57}& {84.12}& {89.46}& {91.73}& {87.09}
\\
{\bf 0.023M} & {Time} & {5.43h} & {0.62h}& {11.45h}& {1.28h}& {1.05h}& {0.72h}& {3.43h}
\\
\bottomrule
    \end{tabular}}
    \caption{The results of DCFT with different convolution kernel sizes were tested on six datasets from GLUE benchmark, and we reported the accuracy and time efficiency. $d$ represents the dimension of the convolutional kernel. We use the same hyperparameters, with only the convolution kernel size varying.}
    \vspace{-0.4cm}
    \label{tab:1}
\end{table*}

\noindent\textbf{Main results}. To evaluate the effectiveness of DCFT fine-tuning for QA tasks, we fine-tuned the DeBERTaV3-base model and assessed its performance on the SQuAD dataset, with results detailed in Table \ref{QA}. We employed identical hyperparameters across our training processes. The outcomes reveal that DCFT surpasses baseline models on two SQuAD datasets, while only necessitating half the parameter count of these baselines. Specifically, on the SQuAD v2.0 benchmark, DCFT achieved scores of 84.7\% and 87.7\% on two key evaluation metrics: exact match (EM) and F1 score, marking an enhancement of 1.1\% and 1.0\% over LoRA.

\subsection{Kernel Sizes Analysis}
\label{size}
For a fixed model, adapting to different downstream tasks is undoubtedly challenging. Some parameters contribute minimally to the final outcomes, not only failing to enhance the model's capabilities but also impacting the convergence speed. Therefore, adaptively adjusting the parameter budget according to needs is crucial. In this section, we have presented the results and training times of DCFT with different convolutional kernel sizes on six different tasks, as shown in Table~\mbox{\ref{tab:1}}. 
The analysis reveals that the model trains fastest and achieves suboptimal overall results when using a kernel size of $(d=8)$. Moreover, when the parameter budget was increased, most tasks showed improved outcomes. When using a kernel size of $(d=2)$, the model achieves the best average results due to the small convolutional kernel's ability to perceive details. However, for the smaller dataset task of CoLA, increasing the parameter budget actually resulted in decreased performance. It is noteworthy that using larger convolutional kernels $(d=12)$ leads to a significant decline in model performance. The primary reason is likely that the enlarged receptive field enables the model to capture excessive irrelevant information, while overlooking critical details, ultimately resulting in deteriorated outcomes.
\subsection{Step Analysis}
In this section, we conduct an ablation experiment on DCFT with different step sizes. We fix the convolution kernel size at eight and keep other hyperparameters consistent. The accuracy (Acc) and training time (Time) for step sizes of $1, 2, 4$, and $8$ across four tasks are recorded. We do not consider cases where the step size exceeds the convolution kernel size, as this would result in parameters not participating in the computation, which contradicts the principles of PEFT. The results are shown in Table \ref{stepacc}. From the analysis, it can be concluded that when the step size equals the kernel size, the model achieves the optimal results with the shortest training time. As the step size decreases, the training time progressively increases. When the step size $= 1$, the model attains the second-best results but incurs nearly three times the training time compared to a step size of 8. We believe the main issue arises when the step size is 4 or 6, as the model is affected by checkerboard artifacts, leading to knowledge overlap and offset. However, with a step size of 1, although the model is still impacted by checkerboard artifacts, the increase in parameter count compensates for this due to enhanced learning capacity, at the expense of significantly prolonged training time. Therefore, the results indicate that setting the step size equal to the convolution kernel size is an effective approach.
\begin{table}[!h]
    \centering
    \scalebox{0.72}{
    \renewcommand\arraystretch{1.2}
    \tabcolsep=0.2cm
    \begin{tabular}{c|c|cccc}
    \toprule
        \textbf{Method\&Params}&\textbf{Metric}&\textbf{CoLA}& \textbf{MRPC}& \textbf{STS-B}&\textbf{RTE}\\
        \midrule
        \textbf{DCFT(step$=1$)}&Acc &70.96&90.69&91.48&\textbf{88.45}\\
        \textbf{165.5K}    &Time&1.99h&1.22h&1.71h&1.62h\\
        \midrule
        \textbf{DCFT(step$=2$)}&Acc &70.04&90.44&91.37&88.09\\
        \textbf{85K}    &Time&0.89h&0.99h&0.89h&1.14h\\
        \midrule
        \textbf{DCFT(step$=4$)}&Acc &65.97&88.97&90.73&87.36\\
        \textbf{44.9k}    &Time&1.03h&0.99h&0.86h&1.11h\\
        \midrule
        \textbf{DCFT(step$=8$)}&Acc &\textbf{71.07}&\textbf{90.93}&\textbf{91.50}&\textbf{88.45}\\
        \textbf{24.6k}    &Time&\textbf{0.54h}&\textbf{0.98h}&\textbf{0.63h}&\textbf{1.08h}\\
    \bottomrule
    \end{tabular}}
    \caption{DCFT uses results with different step sizes, while we adopt the same convolution kernel size $(d=8)$ and ensure that other hyperparameters remain consistent. The best results are highlighted in \textbf{bold}. }
    \vspace{-0.2cm}
    \label{stepacc}
\end{table}

\begin{figure}
\centerline{\includegraphics[width=\linewidth]{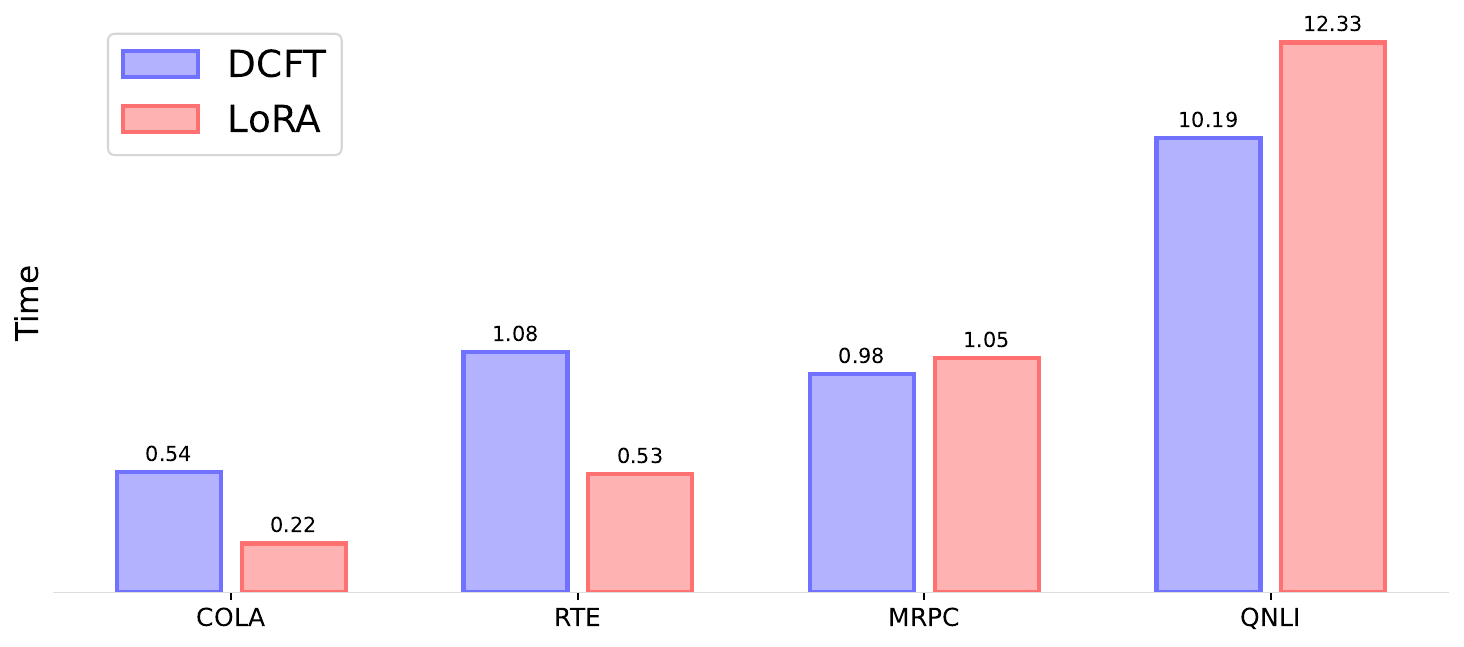}}
  \caption{Illustration of the total training time for DCFT and LoRA on four datasets: COLA, RTE, MRPC, and QNLI.}
  \label{time}
\end{figure}
\begin{figure}
\centerline{\includegraphics[width=\linewidth]{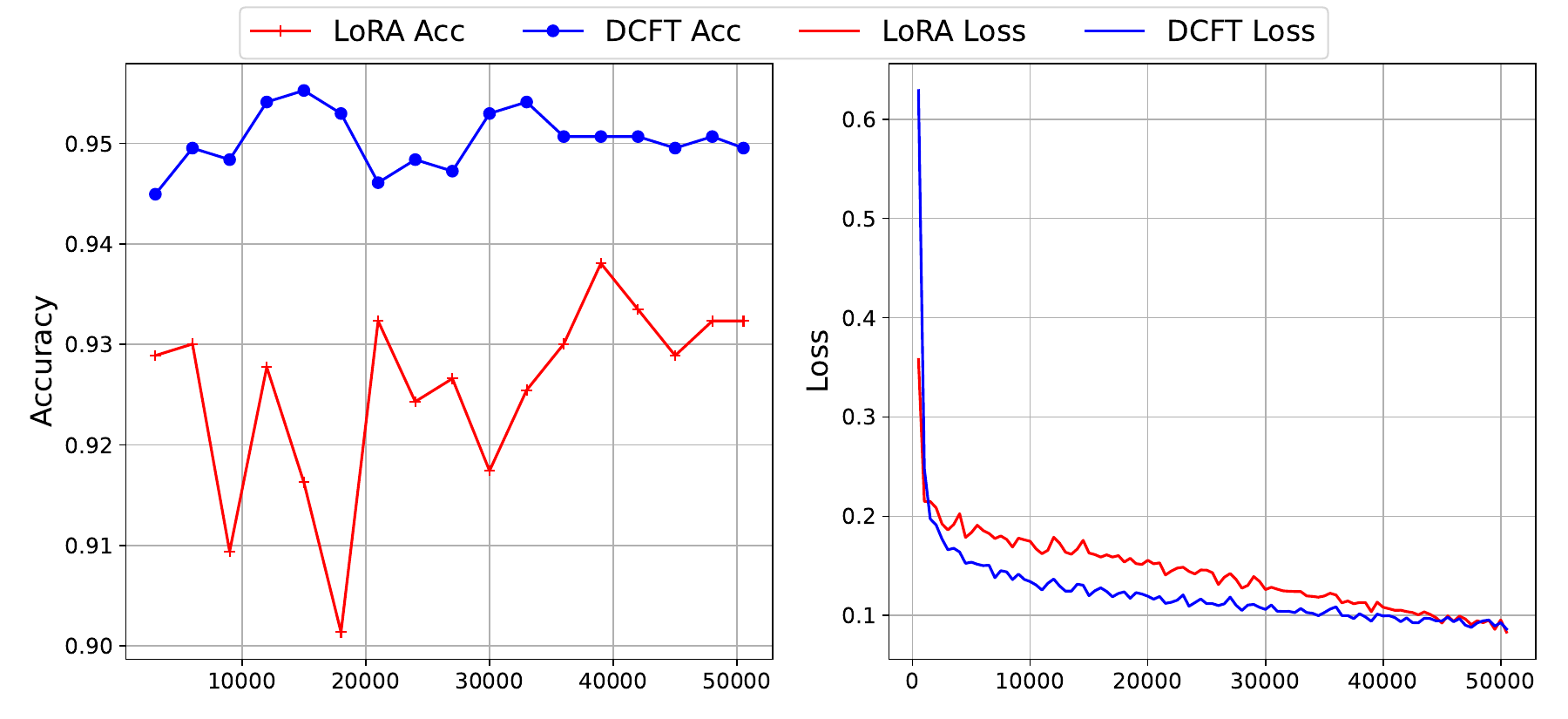}}
  \caption{Accuracy and loss results of DCFT and LoRA on the SST-2 dataset.}
  \label{efficiency}
\end{figure}
\subsection{Efficiency Analysis}
In this section, we compare the training efficiency of DCFT and LoRA. Operating under the same computational infrastructure and with a batch size of 32, we calculate the total training time for four tasks, as shown in Figure~\mbox{\ref{time}}. The results indicate that on smaller tasks like CoLA and RTE, DCFT's training efficiency is lower than LoRA's. However, as the task size increases, DCFT's training efficiency surpasses that of LoRA, particularly on the QNLI task, where it is 17.36\% faster than LoRA.

Analysis shows that due to the high computational overhead of the deconvolution operation, DCFT performs slower than LoRA on small datasets. However, as the size of the task dataset increases, this complexity becomes negligible compared to the overall data processing load, and fewer parameters mean faster checkpoint storage and retrieval. Additionally, the advantage of fewer parameters in DCFT results in significantly reduced gradient computations at each step. Over long periods of computation, this difference accumulates, ultimately leading to a substantial reduction in training time. Furthermore, as shown in Table~\mbox{\ref{time}}, DCFT converges faster than LoRA due to having fewer parameters. Therefore, with the trend towards larger models and fine-tuning datasets, the computational efficiency of DCFT will be further amplified.

\begin{table}[!h]
    \centering
    \scalebox{0.72}{
    \renewcommand\arraystretch{1.2}
    \tabcolsep=0.2cm
    \begin{tabular}{l|c|cccc}
    \toprule
        \textbf{Method}&\textbf{\#Params}&\textbf{CoLA}& \textbf{MRPC}& \textbf{STS-B}&\textbf{RTE}\\
        \midrule
        \textbf{DCFT$_{(Q,K)}$}&6k &64.99&87.25&90.55&83.03\\
        \textbf{DCFT$_{(Q,V)}$}&6k &64.50&89.95&91.43&83.03\\
        \textbf{DCFT$_{(Q,K,V)}$}&9k &65.31&88.97&91.24&84.12\\
        \textbf{DCFT$_{(Q,K,V,O)}$}&18.4k &68.37&89.95&91.50&85.92\\
        \textbf{DCFT$_{(ALL)}$}&24k &\textbf{71.06}& \textbf{90.93}& \textbf{91.56}& \textbf{88.45}\\
    \bottomrule
    \end{tabular}}
    \caption{The results of applying DCFT across various layers are presented. The term $ALL$ denotes the outputs from the query$(Q)$, key$(K)$, value$(V)$, output matrix$(O)$, and feed-forward layers. }
    \vspace{-0.2cm}
    \label{qkv}
\end{table}

\subsection{Applying DCFT to Different layers}
Notably, performance may fluctuate when applying parameter-efficient fine-tuning to different positions within the model. In this section, we apply DCFT to weight matrices at various positions to determine how to achieve optimal performance on downstream tasks. 
Using the DeBERTaV3-base model, we fine-tuned on the CoLA, MRPC, STS-B, and RTE datasets. As shown in Table \ref{qkv}, although DCFT is not primarily designed with parameter budget in mind, applying it to all weight matrices achieves the best results. Applying it specifically to the matrices $Q,K,V,O$ follows closely in effectiveness, achieving the best results on the STS-B dataset. This suggests that, applying DCFT to all weight matrices is effective, and omitting the feed-forward layers result in a significant performance decline.

\section{Conclusion}
We propose a method named DCFT, which innovatively combines the reconstruction capabilities of deconvolution with the learning abilities of incremental matrices. This method enhances and completes the details of knowledge learned in subspaces and allocates parameter budgets based on the size of convolution kernels, free from the constraints of rank one decomposition. Additionally, we have implemented three principal measures to reduce the computational complexity of DCFT. This effectively lowers the model's computational complexity, allowing it to achieve better training efficiency on large datasets compared to LoRA. We conducted extensive experiments in natural language processing and question answering. The results show that DCFT achieves improved performance while reducing parameters by approximately 8$\times$.
% \section*{Limitations}
% Although LoRA$^2$ has demonstrated surprising performance on NLP datasets, our research still has some acknowledged limitations. However, recent studies have shown that methods for parameter-efficient fine-tuning can also be applied in the cross-modal domain, and the performance of LoRA$^2$ in the multimodal field is currently unknown. Additionally, our method only evaluates the fine-tuning results of dual LoRA at multiple scales. For more multi-scale LoRA, we intend to conduct experiments to verify its performance.
\section*{Acknowledgements}
This work was supported in part by the Science and Technology Commission of Shanghai Municipality under Grant (21DZ2203100), in part by the National Natural Science Foundation of China under Grant (62006150), in part by Shanghai Local Capacity Enhancement project (21010501500) and in part by Science and Technology Innovation Action Plan of Shanghai Science and Technology Commission for social development project under Grant (21DZ1204900).

% \newpage
\bibliography{custom}
\clearpage
\newpage
\appendix
\section{Datasets}
\subsection{NLU Datasets}
\label{A}

For evaluation, we adaopt the GLUE benchmark \cite{wang-etal-2018-GLUE}, including CoLA \cite{warstadt-etal-2019-neural}, SST-2 \cite{socher-etal-2013-recursive}, MRPC \cite{dolan-brockett-2005-automatically}, QQP \cite{wang-etal-2018-GLUE}, STS-B \cite{wang-etal-2018-GLUE}, MNLI \cite{williams-etal-2018-broad}, QNLI \cite{rajpurkar-etal-2016-squad} and RTE \cite{10.1007/11736790_9,giampiccolo-etal-2007-third,Bentivogli2011TheSP}. We present the dataset statistics of GLUE in the following table \ref{tab:NLU}.

\begin{table}[!h]
    \centering
    \scalebox{0.77}{ 
    \renewcommand\arraystretch{1.5}
    \begin{tabular}{l|ccccc} 
    \toprule
      \textbf{Dataset} &  \textbf{Metric}  &  \textbf{\#Train} & \textbf{\#Valid} & \textbf{\#Test}& \textbf{\#Label}  \\ \midrule
        CoLA & Mcc&8.5k &1,043 &1,063 &2  \\
        SST-2& Acc &67k &872 &1.8k &2 \\
        MRPC & Acc&3.7k &408 &1.7k &2\\
        QQP  & Acc/F1&364k &40.4k &391k &2 \\
        STS-B& Corr&5.7k &1.5k &1.4k &1 \\
        MNLI & Acc(m/mm)&393k &20k &20k&3 \\
        QNLI & Acc&105k &5.5k &5.5k &2 \\
        RTE  & Acc&2.5k &277 &3k &2\\
    \bottomrule
    \end{tabular}}
    \caption{ Dataset Sizes and Evaluation Metrics in the GLUE Benchmark. "Mcc," "Acc," "F1," and "Corr" denote the Matthews correlation coefficient, accuracy, F1 score, and Pearson correlation coefficient, respectively. "Acc(m/mm)" indicates accuracy results for matched and mismatched datasets within MNLI.}
    \label{tab:NLU}
\end{table}

\subsection{QA Datasets}
\label{B}

For evaluation, we adaopt (QA) datasets: SQuAD v1.1 \cite{rajpurkar-etal-2016-squad} and SQuAD v2.0 \cite{rajpurkar-etal-2018-know}. We present the dataset statistics of GLUE in the following table \ref{tab:QA}.

\begin{table}[!h]
    \centering
    \scalebox{0.85}{ 
    \tabcolsep=0.4cm
    \renewcommand\arraystretch{1.6}
    \begin{tabular}{l|ccc} 
    \toprule
      \textbf{Dataset} &  \textbf{Metric}  &  \textbf{\#Train} & \textbf{\#Valid} \\ \midrule
        SQuAD v1.1 & EM/F1 &87,599   &10,570 \\
        SQuAD v2.0 & EM/F1 &130,319  &11,873 \\
    \bottomrule
    \end{tabular}}
    \caption{ Dataset Sizes and Evaluation Metrics in the SQuAD Benchmark. "EM," and "F1," denote the Exact Match and F1 Score.}
    \label{tab:QA}
\end{table}

% \subsection{NLG Datasets}
% \label{C}
% We fine-tune Bart-large \cite{lewis-etal-2020-bart} and utilize the XSum \cite{xsum-emnlp} and CNN/DailyMail \cite{hermann2015teaching} datasets for evaluation. XSum and CNN/DailyMail are both used for text summarization tasks, the former provides single-sentence summaries, while the latter offers multi-sentence summaries paired with news articles. We present the dataset statistics of GLUE in the following table \ref{tab:NLG}.

% \begin{table}[!h]
%     \centering
%     \scalebox{0.85}{ 
%     \renewcommand\arraystretch{1.6}
%     \begin{tabular}{l|cccc} 
%     \toprule
%       \textbf{Dataset} &  \textbf{Metric}  &  \textbf{\#Train} & \textbf{\#Valid} & \textbf{\#test}\\ \midrule
%         XSum      & R-1/2/L & 204,045   & 11,332 & 11,334 \\
%         CNN       & R-1/2/L & 90,266    & 1,220  & 1,093 \\
%         DailyMail & R-1/2/L & 196,961   & 12,148 & 10,397 \\
%     \bottomrule
%     \end{tabular}}
%     \caption{ Dataset Sizes and Evaluation Metrics in the SQuAD Benchmark. "EM," and "F1," denote the Exact Match and F1 Score.}
%     \label{tab:NLG}
% \end{table}

\section{Sparse Regularization Theory}

By using progressive projection matrices, we further increase the compression ratio $p$ of the parameters. Additionally, this enhances the stability of the fine-tuning process. equations \ref{eq:15}\cite{10.1609/aaai.v37i11.26505} theoretically demonstrate the role of sparsity in model stability. As the compression ratio $p$ decreases, the upper bound also decreases. Therefore, a sparser model implies better stability.
\begin{align}
\label{eq:15}
 E_{S,i\sim U(n)}[|\ell(A(S^{i}),z_i)-\ell(A(S),z_i)|]\nonumber& \\
 \le \frac{2\rho^2 }{(\Lambda_{min}+2(1-p))n},
\end{align}\\
$E_{S,i\sim U(n)}[\cdot]$ is Pointwise Hypothesis Stability (PHS) \cite{10.1145/567806.567809}  which focuses on analyzing the change of model output after a training sample is removed. $\Lambda_{\text{min}} = \min\{\Lambda_1, \dots, \Lambda_m\}$. $\ell(\cdot)$ represents the loss function. The variable $\rho$ represents this measure of stability, reflecting the maximum impact of input variations on the output in the loss function. 

\begin{table*}[htbp]
    \centering
    \scalebox{0.80}{
    \tabcolsep=0.5cm
    \renewcommand\arraystretch{1.2}
    \begin{tabular}{l|cccccccc}
    \toprule
    \textbf{Task}&learning rate& batch size & $r$& Epochs&Dropout & $\gamma$ &init\_warmup &$\Delta_{T}$ \\  
        \midrule
        \textbf{CoLA} &5e-4  &32 &8 &25 &0.5 &0.1 &800   &10  \\ 
        \textbf{SST-2}&8e-4  &32 &8 &24 &0.1 &0   &6,000 &100 \\
        \textbf{MRPC} &1e-3  &32 &8 &30 &0.1 &0   &600   &1   \\
        \textbf{QQP}  &8e-4  &32 &8 &20 &0.1 &0.15&8,000 &100 \\
        \textbf{STS-B}&2.2e-3&32 &8 &25 &0.1 &0.2 &800   &10  \\
        \textbf{MNLI} &5e-4  &32 &8 &7  &0.1 &0.15&8,000 &100 \\
        \textbf{QNLI} &1.2e-3&32 &8 &6  &0.1 &0.1 &2,000 &100 \\
        \textbf{RTE}  &1.2e-3&32 &8 &50 &0.3 &0.2 &600   &1   \\
    \bottomrule
    \end{tabular}}
    \caption{Hyper-parameters setup of DCFT for GLUE benchmark.}
    \vspace{-0.4cm}
    \label{HyperparametersNLU}
\end{table*}
\begin{table*}[htbp]
    \centering
    \scalebox{0.80}{
    \tabcolsep=0.45cm
    \renewcommand\arraystretch{1.2}
    \begin{tabular}{l|cccccccc}
    \toprule
    \textbf{Task}&learning rate& batch size & $r$& Epochs& $\gamma$ &init\_warmup &$\Delta_{T}$ &$t_{f}$ \\  
        \midrule
        \textbf{AQuAD v1.1} &1e-3  &16 &10 &25 &0.1 &5,000  &100 &25000 \\ 
        \textbf{AQuAD v2.0} &1e-3  &16 &12 &24 &0.1 &5,000  &100 &50000\\
    \bottomrule
    \end{tabular}}
    \caption{Hyper-parameters setup of DCFT for AQuAD benchmark.}
    \vspace{-0.4cm}
    \label{HyperparametersQA}
\end{table*}
% \begin{table*}[!]
%     \centering
%     \scalebox{0.80}{
%     \renewcommand\arraystretch{1.2}
%     \begin{tabular}{l|ccccccc}
%     \toprule
%     \textbf{Task}&learning rate& batch size & Epochs& $\gamma$ &init\_warmup &$\Delta_{T}$&$t_{f}$ \\  
%         \midrule
%         \textbf{XSum}           &5e-4 &64   &25  &0.1 &6000 &100 &50000\\
%         \textbf{CNN/DailyMail}  &5e-4 &32   &15  &0.1 &5000 &100 &85000\\
%     \bottomrule
%     \end{tabular}}
%     \caption{Hyper-parameters setup of DCFT for NLG tasks.}
%     \vspace{-0.4cm}
%     \label{HyperparametersNLG}
% \end{table*}
\section{Orthogonal Projection Theory}
It is a fundamental concept in linear algebra with applications across various fields including machine learning, statistics, and computer graphics. This theory revolves around the idea of projecting a vector onto a subspace in a way that minimizes the distance between the vector and the subspace, effectively finding the closest approximation within that subspace.

Mathematically, consider a vector $u$ in $R_n$ and a subspace $\mathbb{V}$ spanned by vectors $ \{v_1, v_2, \ldots, v_k\} $. The orthogonal projection of u onto $\mathbb{V}$, denoted as $\mathbb{P}_\mathbb{V}(\mathbf{u})$, is given by:
\begin{equation}
\label{eq:23}
\mathbb{P}_\mathbb{V}(\mathbf{u}) = \sum_{i=1}^k \frac{\mathbf{u} \cdot \mathbf{v}_i}{\mathbf{v}_i \cdot \mathbf{v}_i} \mathbf{v}_i.
\end{equation}
This design allows each Low Rank block to capture information in different dimensions, thereby reducing information overlap and increasing the overall efficiency and effectiveness of the model. Additionally, the orthogonal training strategy helps prevent overfitting, making the model more robust when faced with new data.

\section{Checkerboard Artifacts}
Checkerboard artifacts, commonly encountered in image generation or upsampling tasks using Convolutional Neural Networks (CNNs), are often associated with the use of strided or transposed convolutions. These operations can lead to uneven coverage in the output images, resulting in visible grid-like patterns.

The formation of checkerboard artifacts can be described by the following equations. Suppose $f(x,y)$ represents the generated image and $k(x,y)$ is the convolutional kernel, applied with stride $s$ during upsampling. Each pixel $p(x,y)$ in the generated image can be obtained through the convolution operation:
\begin{equation}
\label{eq:24}
p(x, y) = \sum_{i, j} f(i, j) \cdot k(x - si, y - sj),
\end{equation}
where $i$ and $j$ traverse all valid pixel coordinates. When the stride $s$ is greater than $1$, $k(x,y)$ may not cover some positions in $x$ or $y$ completely, causing discontinuities at these points in the output $p(x,y)$ and forming a checkerboard pattern.

To mitigate this issue, researchers have developed various strategies, such as replacing strided convolutions with interpolation methods (like bilinear interpolation) or optimizing the design of convolution kernels to avoid uneven overlaps. Another approach involves adjusting the relationship between stride and kernel size to ensure uniform influence on each output pixel. For instance, fractionally strided convolutions can be utilized, where the formula is modified to:
\begin{equation}
\label{eq:25}
p(x, y) = \sum_{i, j} f(i, j) \cdot k\left(\frac{x}{s} - i, \frac{y}{s} - j\right).
\end{equation}
Through these methods, checkerboard artifacts can be reduced or eliminated to enhance image quality.
\section{Hyperparameters}
\label{E}

Regarding hyperparameters, We tune the learning rate and pick the best learning rate for every method. For each dataset, the batch size is set as identical for every method. The majority of hyperparameters used in evaluating DCFT on Natural Language Understanding and Question Answering are shown in Table \ref{HyperparametersNLU} and \ref{HyperparametersQA}.

% \ref{HyperparametersNLG}.

\end{document}